\documentclass{ecai}

\usepackage{latexsym}
\usepackage{amssymb}
\usepackage{amsmath}
\usepackage{amsthm}
\usepackage{booktabs}
\usepackage{enumitem}
\usepackage{graphicx}
\usepackage{color}
\usepackage{xcolor}

\makeatletter
\let\c@author\relax
\makeatother

\usepackage[style=numeric-comp,maxnames=3,natbib=true,firstinits=true]{biblatex}
\DeclareSourcemap{
  \maps[datatype=bibtex, overwrite]{
    \map{
        \step[fieldset=editor, null]
      \step[fieldset=publisher, null]
      \step[fieldset=volume, null]
      \step[fieldset=series, null]
    }
  }
}

\newcommand{\BibTeX}{B\kern-.05em{\sc i\kern-.025em b}\kern-.08em\TeX}

\newcommand{\reb}[1]{\textcolor{black}{#1}}
\newcommand{\ours}{ExpertSim}

\usepackage{hyperref}
\usepackage{caption}
\usepackage{ragged2e}
\begin{document}

\begin{frontmatter}

\paperid{1471}

\title{ExpertSim: Fast Particle Detector Simulation \\ Using Mixture-of-Generative-Experts}

\author[A]{\fnms{Patryk}~\snm{Będkowski}}
\author[A,B]{\fnms{Jan}~\snm{Dubiński}}
\author[A,C]{\fnms{Filip}~\snm{Szatkowski}} 
\author[A,D]{\fnms{\\Kamil}~\snm{Deja}}
\author[A]{\fnms{Przemysław}~\snm{Rokita}}
\author[A,D]{\fnms{Tomasz}~\snm{Trzciński}}

\address[A]{Warsaw University of Technology}
\address[B]{NASK National Research Institute}
\address[C]{IDEAS NCBR}
\address[D]{IDEAS Research Institute}

\begin{abstract}
Simulating detector responses is a crucial part of understanding the inner workings of particle collisions in the Large Hadron Collider at CERN. Such simulations are currently performed with statistical Monte Carlo methods, which are computationally expensive and put a significant strain on CERN's computational grid. Therefore, recent proposals advocate for generative machine learning methods to enable more efficient simulations. However, the distribution of the data varies significantly across the simulations, which is hard to capture with out-of-the-box methods. In this study, we present ExpertSim - a deep learning simulation approach tailored for the Zero Degree Calorimeter in the ALICE experiment. Our method utilizes a Mixture-of-Generative-Experts architecture, where each expert specializes in simulating a different subset of the data. This allows for a more precise and efficient generation process, as each expert focuses on a specific aspect of the calorimeter response. ExpertSim not only improves accuracy, but also provides a significant speedup compared to the traditional Monte-Carlo methods, offering a promising solution for high-efficiency detector simulations in particle physics experiments at CERN. We make the code available at \url{https://github.com/patrick-bedkowski/expertsim-mix-of-generative-experts}. %
\end{abstract}

\end{frontmatter}

\section{Introduction}

ALICE~(A Large Ion Collider Experiment) is one of the four major detectors located at the Large Hadron Collider~(LHC) at CERN. 
One of its main goals is to replicate and study the intense conditions that existed in the early universe shortly after the Big Bang. 
To understand the physics behind this state, LHC collides particles accelerated almost to the speed of light, and gather the data describing the effect of those collisions. However, to validate hypotheses, collected data needs to be statistically compared with theoretical simulations.
Such simulations are extremely computationally expensive, as existing approaches utilize statistical Monte-Carlo methods to model the physical interactions between particles. While these methods yield high-fidelity outcomes, they are also associated with high computational demands. In 2023, over 540 000 CPU devices \citep{lhccomputationreport} were engaged in the computations of ALICE experiments, marking a demand for developing more efficient simulation techniques.

The high computational cost of the simulations calls for alternative, lightweight solutions to accelerate the experimentation cycle. Throughout recent years, generative deep neural networks have emerged as a promising alternative, that does not require repetitive calculations present in the Monte Carlo approach. Thanks to that, recent works show over tens or hundred times faster simulations~\citep{deja2018generative,deja2020endtoend}, with possible straightforward parallelization on high performance GPUs. 
However, applying standard machine learning algorithms to this domain requires significant modifications, as the data distributions in physical experiments do not resemble the traditional distributions encountered \textit{e.g.} in computer vision.
Therefore, maintaining the high fidelity and diversity of the simulations with generative models remains a challenging task due to the high variance in the data distribution.

This is especially true for the most computationally expensive simulation of the Zero Degree Calorimeter~(ZDC).
This device is designed to measure proton energy in heavy ion collisions and plays a crucial role in monitoring the centrality of particle collisions at the ALICE experiment. By design, most of the ZDC responses fall into one of the three general groups that exhibit significantly different properties.
Recent techniques in this domain tackled the problem of simulating ZDC's responses ~\citep{Dubinski2023MachineLM,sdigan2023} aiming to increase the simulation quality without compromising the speed. However, modeling multiple distinct distributions with a single model requires high model capacity and complexity, which is contrary to the goal of improving the simulation speed. In this study, we propose to solve this problem through a 
Mixture-of-Experts~(MoE) approach that allows us to maintain the fast generation speed without compromising the simulation quality.

MoE layers~\citep{shazeer2016outrageously} are composed of multiple expert modules and a gating network. The gating network determines the information flow in the network by selecting which experts are utilized for a given input. MoE leverages the fact that different inputs to the neural networks might require different treatments, which allows for more efficient processing. Inspired by MoE methods, we propose to utilize a similar approach for the ZDC data simulation. To that end, we introduce a single generative MoE model that dynamically selects small, specialized expert for a given input to enable fast processing with high fidelity.
In our approach, each expert is based on the deep convolutional generative adversarial network (GAN)~\citep{gan}. To increase the diversity of generated samples within each expert, we use SDI-GAN model~\citep{sdigan2023} architecture that adds a diversity regularization term to the standard GAN objective.
Experts are assigned specific samples to process by an efficient routing network, which we train leveraging specific physical characteristics of the samples.
Following \cite {bedkowski2024deep}, we also employ a regularization method focused on minimizing the difference in intensities between real and generated calorimeter responses, which improves the quality of the simulations. Finally, we introduce an auxiliary regressor which increases the model capabilities to learn accurate spatial features of the simulation data.

\ours{} significantly outperforms all existing approaches and achieves the highest simulation fidelity while maintaining nearly the same inference time as single-model methods. 
We improve over the previous state-of-the-art by more than 15\% and achieve a Wasserstein distance of 1.70 between the distributions of real and generated data.
Our approach offers a promising avenue towards providing fast and reliable simulations in a challenging CERN environment.

We summarise the contributions of this work as follows:
\begin{itemize}
    \item We propose \ours{}, a novel, MoE-based generative model architecture simulating Zero Degree Calorimeter responses in the ALICE experiment at CERN.
    \item We introduce a router training scheme that leverages the physical properties of the data, which results in high expert specialization and precise routing decisions. 
    \item We evaluate our simulation model in real-life scenarios and compare it with single model approaches, demonstrating over 15\% improvement upon the previous state-of-the-art without compromising the generation speed.
\end{itemize}

\section{Related work}

\subsection{Generative models for fast simulation in CERN}

Throughout recent years, the use of Generative AI for various CERN simulations has shown the versatility of these methods. The majority of works focus on the usage of generative adversarial networks~\citep{paganini2018calogan, sofia18, di2019dijetgan, erdmann2019precise}. Alternative approaches include generative autoencoders~\citep{deja2020endtoend, howard2022learning, rogoziński2024particlephysicsdlsimulationcontrol} or more recently diffusion models~\citep{cresswell2022caloman, kita2024generativediffusionmodelsfast} and normalizing flows~\citep{wojnar2024applying, wojnar2025fast, wojnar2025faster}.

For the particular case of ZDC, there are several approaches based on GANs~\citep{Dubinski2023MachineLM,sdigan2023} or Sinkhorn autoencoder~\citep{deja2020endtoend}.  
In particular, \citep{Dubinski2023MachineLM} employs generative machine learning algorithms for the task of simulating a Neutron ZDC. They propose a solution that utilizes variational autoencoders~\citep{vae} and generative adversarial networks~\citep{gan}. By expanding the GAN architecture with an additional regularization, the authors significantly increase the simulation speed by two orders of magnitude while maintaining the high simulation fidelity.

In this vein of research, to match the diversity of the simulation observed in the training data, in \citep{sdigan2023} authors propose a GAN model dubbed SDI-GAN. The method forces the generator to discover new data modes for inputs related to diverse outputs while generating consistent samples for the remaining ones.

\subsection{Mixture-of-Experts}

Many methods have been crafted to reduce the computational cost of training and inference of deep neural networks, including network architectures designed for efficiency~\citep{tan2019efficientnet}, knowledge distillation~\citep{hinton2015distilling}, pruning and quantization~\citep{liang2021pruning}. Additionally, multiple works also try to reduce resource usage through the introduction of conditional computations, leveraging the fact that different data points might require different conditional complexity to save compute on the easy samples. Some works adapt the compute by selecting the subset of filters~\citep{liu2019learning,herrmann2020channel}, features~\citep{figurnov2017spatially,verelst2020dynamic} or tokens~\citep{riquelme2021scaling} processed in each layer. Works such as \citep{lin2017runtime,nie2021evomoe} introduce sparsity while training the model. Multiple methods~\citep{graves2016adaptive,wang2018skipnet,dehghani2018universal,elbayad2019depth,banino2021pondernet} allow the model to adapt its depth to the input example via skipping layers~\citep{wang2018skipnet}, competing halting scores~\citep{graves2016adaptive, banino2021pondernet}, introducing recursive computations~\citep{dehghani2018universal,elbayad2019depth} or attaching early exit classifiers~\citep{scardapane2020should,matsubara2022split,wojcik2023zero}. However, perhaps the most widely spread dynamic network architecture is Mixture-of-Experts~(MoE). 

MoE was introduced as an efficient way to further increase the capacity of deep neural networks applied in NLP, initially in LSTM models~\citep{shazeer2016outrageously}, and later in Transformers~\citep{lepikhin2020gshard}. Since then, they have also been adapted to computer vision~\citep{riquelme2021scaling,daxberger2023mobile}. MoE layers have gained significant popularity primarily due to their excellent scaling properties~\citep{du2022glam,clark2022unified}. Nonetheless, training such models is challenging, primarily because gating decisions must be discrete to ensure sparse expert selection. Various methods of training were proposed, some of which include reinforcement learning \citep{bengio2013estimating}, weighting the expert output by the probability to allow computation of the gradient of the router~\citep{shazeer2016outrageously}, or using the Sinkhorn algorithm~\citep{clark2022unified}. Some of those approaches also suffer from the possibility of load imbalance and therefore require auxiliary losses or alternative expert selection methods~\citep{fedus2022switch,zhou2022mixture}. Interestingly, in many cases, fixed routing functions perform similarly to trainable routers~\citep{roller2021hash}, suggesting that current solutions are largely suboptimal. MoE models can also be derived from pre-trained dense models by splitting the model weights into experts and independently training the routers for each layer~\citep{zhang2022moefication,zuo2022moebert}, which avoids most of the problems present in end-to-end training.

Other works explore utilizing experts for different data subsets and introduce domain experts~\citep{gururangan-demix, krishnamurthy2023improvingexpertspecializationmixture}, task-specialized~\citep{Chen_2023_CVPR} or finely-grained experts~\citep{dai2024deepseekmoe} designed to maximize the specialization. In the context of generative models, similar strategies utilizing experts to cover different data subsets were developed for GANs~\citep{park2018megan} and multi-modal autoencoders~\citep{shi2019variational}. 

The key novelty of \ours{} with respect to existing approaches lies in leveraging and extending MoE ideas to address a real-world problem of fast, high-fidelity simulations of particle collision. Unlike standard transformer-based MoEs, where increasing parameter count is a primary goal, our motivation for using multiple experts is to enable the model to capture the diverse range of detector responses observed in practice, which cannot be effectively modeled by a single generative networks. Furthermore, the high-energy physics domain’s need for fast simulations makes GANs the preferred modeling backbone for each expert. This is a significant distinction from conventional MoEs, where experts are typically implemented as multilayer perceptrons. 

\section{Zero Degree Calorimeter simulation}

The Zero Degree Calorimeter (ZDC) \citep{zdc_alice} includes a Proton (ZP) and a Neutron (ZN) device for recording energy from non-interacting nuclei in collisions. The role of this device is to provide information about the centrality of particle collision undergoing inside the ALICE experiment. The higher the energy of non-interacting nuclei measured by ZDC, the lower the centrality of the recorded collisions. Both ZP and ZN are quartz-fiber calorimeters, utilizing the principle of detecting Cherenkov light produced by charged particles of the shower in silica optical fibers \citep{zdc_alice}. Every alternate fiber is directed towards a photomultiplier (PMTc), with the rest of the fibers being grouped into bundles that lead to four distinct photomultipliers (PMT1 to PMT4). The design allows for precise measurement of particle energies in heavy ion collisions at the CERN LHC, with ZP and ZN capturing distinct particle types.

As the silica fibers are arranged in a $n$ by $n$ grid, we treat the response of the ZDC as a 1-channel  $n\times n$ image with pixel values equal to the number of photons deposited in the respective optical fiber. Each simulation example is, therefore, an image coming from a ZP or ZN device, which is referred to as a response of the experiment associated with a vector of variables, referred to as conditional data. Conditional data compromises of 9 variables: energy, mass, charge, three spatial position coordinates, and three momentum coordinates. The resolution of ZP and ZN responses is $56\times30$ and $44\times44$ respectively.

\begin{figure*}[h]
    \centering
    \includegraphics[width=0.8\textwidth]{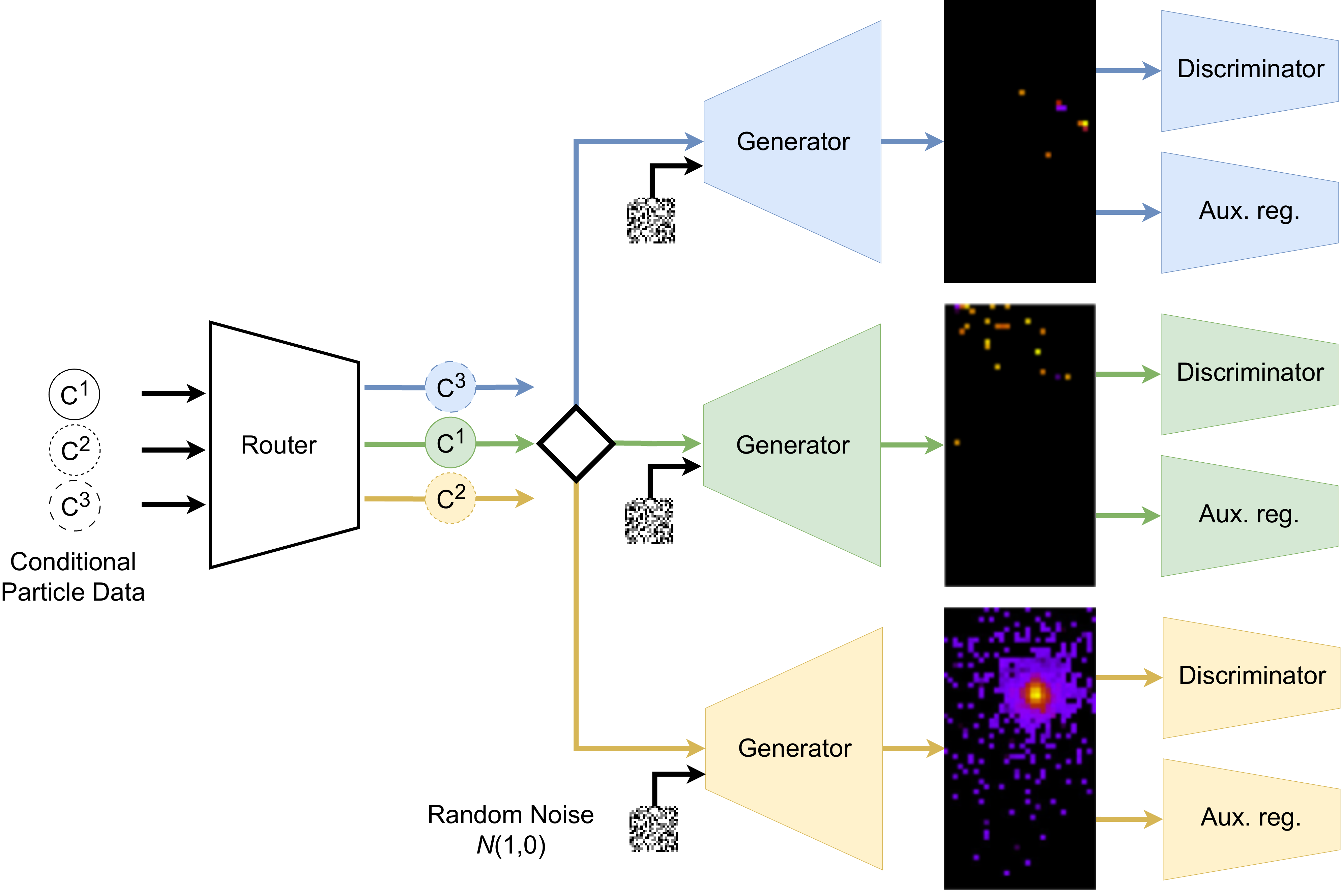}
    \vspace{0.2cm}
    \caption{Overview of \ours{} architecture. Our approach leverages the Mixture-of-Experts paradigm to dynamically route particle data to one of the three specialized generators, which enables us to use a small-sized generator while maintaining the generation quality.}
    \label{fig:model_overview}
    \vspace{0.2cm}
\end{figure*}

\section{Method}

Simulating the responses of the ZDC at ALICE CERN requires capturing the complex relationships between particle properties and calorimeter outputs. These relationships are often too intricate for a single generative model. To address this challenge, we propose \ours{}, a framework comprising a mixture of generative experts presented in Fig.~\ref{fig:model_overview}.

Specifically, \ours{} is composed of a router network and three generative expert models. The router is a fully connected neural network that takes conditional particle data as input. We train the router to optimize for specialization, guiding it to allocate particles that produce similar outputs to the same generative expert.

Each generative expert in our framework is implemented as a generative adversarial network (GAN) consisting of a Generator, an Auxiliary Regressor, and a Discriminator. The Generator takes random noise from a normal distribution, \(N(0,1)\), conditioned on particle data to synthesize realistic images. The Auxiliary Regressor enhances spatial learning by estimating the coordinates of the image’s high-intensity region, aiding in the geometric accuracy of generated outputs. The Discriminator, also conditioned on particle data, distinguishes between real and synthetic images, refining the Generator’s outputs through adversarial training. This multi-expert setup enables \ours{} to more effectively capture the nuanced response patterns of the ZDC, offering improved accuracy and specialization over traditional single-model generative approaches.

\subsection{Expert generative adversarial networks}
As described above, we employ a Deep Convolutional generative adversarial network (DCGAN)~\citep{gan} as a single generative expert, which consists of a generator, $G(z, c)$, and a discriminator, $D(x, c)$. The generator synthesizes an image $x$ starting from random noise $z$, while the discriminator attempts to distinguish between real and generated images. Both the generator and discriminator are conditioned on particle data $c$ and trained in an adversarial manner. After training, the generator is used to create new calorimeter response images. %
Formally, given a conditional input $c$ and a random noise $z$ sampled from $k$ independent standard normal distributions, $z \sim \mathcal{N}(0,1)^k$, the generator $G(z, c)$ produces an output image $\hat{x} = G(z, c)$.

\subsubsection{Diversity Regularization}

SDI-GAN~\citep{sdigan2023} introduces a regularization approach to encourage diversity by minimizing the inverted ratio of the $L1$ distance between two generated images $d_I$, produced from distinct latent codes $z_1$ and $z_2$, and the $L1$ distance between these codes $d_z$, under the same conditioning vector $c$. The diversity metric relies on the pixel variance observed in the original dataset to scale the regularization term, accounting for varying levels of diversity across different conditioning inputs. For each unique conditioning value $c$, the variance of pixel values across samples is calculated during the preprocessing stage. The diversity values for each sample are normalized to a $[0, 1]$ range by dividing by the dataset size. This measure is then scaled by the regularization term $\lambda_{div}$, adjusting its strength during training:
\begin{equation}
L_{div} = \sum_{i, j} \sqrt{\frac{\sum_{t}(x_{ij}^{t} - \mu_{ij})^2}{|X|}} \times \left(\frac{d_I (G(c, z_1), G(c, z_2))}{d_z (z_1, z_2)} \right)^{-1},
\end{equation}
where $i$ and $j$ represent pixel coordinates, $t$ is the index for a sample $x \in \chi$, and $\mu_{ij}$ is the mean pixel value for coordinates $(i, j)$.

\subsubsection{Intensity Regularization}

While SDI-GAN performs well on filtered data, it can struggle with variations in Cherenkov light intensity across distributions. To address this, we introduce a regularization term based on an intensity measure $f_{in}$ derived from the original dataset. During preprocessing, the intensity measure for each conditioning vector $c$ is calculated as the sum of pixel values in the corresponding image $x$ from $\chi$:
\begin{equation}
    \label{eq:intensity}
    f_{in}(x) = \sum_{i, j}x_{ij}, 
\end{equation}
where $i$ and $j$ denote pixel coordinates. The intensity difference between a generated image $\hat{x}$ and the corresponding real image $x$ is computed with Mean Absolute Error~(MAE). The loss is also weighted by a constant $\lambda_{in}$ to control its impact:
\begin{equation}
L_{in} = |f_{in}(x_c) - f_{in}(\hat{x}_c)|.
\end{equation}

\subsubsection{Auxiliary Regressor}

A key aspect of calorimeter response is identifying the location of the shower center, where pixel intensities are the highest. To help the individual experts better capture these geometric properties, we integrate an auxiliary regressor that works alongside the main model. This regressor is tasked with pinpointing the 2D coordinates of the collision center. During preprocessing, these coordinates are determined for all training samples and used as targets for the regressor. The auxiliary loss is calculated as the mean squared error (MSE) between the predicted coordinates $(\hat{k_{i}}, \hat{l_{i}})$ and the actual coordinates $(k_{i}, l_{i})$ of the peak-intensity pixel in the image $x_i$. This loss is then added to the generator's overall loss function to improve its geometric accuracy, with its influence controlled by the parameter $\lambda_{aux}$:
\begin{equation}
L_{aux} = \frac{1}{N} \sum_{i=1}^{N} \left[ (\hat{k_{i}} - k_{i})^2 + (\hat{l_{i}} - l_{i})^2 \right].
\end{equation}

\subsubsection{Final GAN Expert Training Objective}

The overall training loss incorporates the traditional GAN loss along with the additional terms introduced above, resulting in the following composite objective:
\begin{align}
L(G, D) &= L_{adv}(G, D) + \lambda_{div}L_{div}(G) \\
&+ \lambda_{in}L_{in}(G) + \lambda_{aux}L_{aux}. \nonumber 
\end{align}
This final loss function blends adversarial loss with diversity, intensity, and auxiliary regression components, each scaled by their respective weights $\lambda_{div}$, $\lambda_{in}$, and $\lambda_{aux}$, to balance the different training objectives.

To ensure a fair comparison between methods, we train all models using identical parameter settings, including the learning rates of the generator, discriminator, and auxiliary regressor, as well as consistent loss strengths across all models. 

\subsection{Router}

The router model is implemented as a multilayer, fully connected neural network that takes particle properties as input and returns an assignment to one of three generative experts. The purpose of the router is to dynamically allocate inputs to the most suitable expert based on the underlying characteristics of the data. By leveraging conditional information on particle properties, the router determines which expert should generate the calorimeter response for each input. This process leads to specialization of experts on different subsets of input data. To achieve this task, we train the router to balance the workload among experts while also promoting their specialization. We achieve this goal with two losses described in the following sections. To balance the load between experts we minimize the entropy-based expert utilization loss. At the same time, to promote their specialization, we introduce an expert differentiation loss that maximizes diversity between the mean outputs of all experts. The combined use of expert utilization and differentiation losses ensures that the router effectively balances task distribution while fostering distinctive behaviors among experts. 

\subsubsection{Expert Utilization Loss}
The expert utilization entropy loss \( L_{\text{util}} \) is calculated as follows:
\begin{equation}
L_{\text{util}} = - \sum_{i=1}^{N} \bar{p}_i \log(\bar{p}_i + \epsilon),
\end{equation}
where $N$ is the number of experts, $\bar{p}_i$ is the average gating probability for expert \( i \), computed over the batch, and $\epsilon$ is a small constant added for numerical stability.

This loss encourages the router to distribute samples evenly across all experts. By maximizing the entropy, the router avoids over-reliance on any single expert, ensuring a balanced load that enhances the model’s ability to specialize and handle diverse inputs effectively.

\subsubsection{Expert Differentiation Loss}
The differentiation loss \( L_{\text{diff}} \) is calculated based on the mean intensities of images generated for each expert. 
The intensity \( f_{\text{in}}(x) \) for an image \( x \) is defined as the sum of pixel values across the image as in the Equation \ref{eq:intensity}. Mean intensities \( \bar{f}_{\text{in}}^{(i)} \) for the generated images corresponding to each expert \( i \) are then used to compute the differentiation loss $L_{\text{diff}}$:
\begin{equation}
    \label{eq:utilization}
L_{\text{diff}} = -\sum_{i=1}^{N} \sum_{j=i+1}^N (\bar{f}_{\text{in}}^{(i)} - \bar{f}_{\text{in}}^{(j)})^2.
\end{equation}

The purpose of the differentiation loss is to encourage diversity among the experts by penalizing cases where the mean intensities of the outputs are similar. The router fosters specialization among experts by maximizing these differences, enhancing its capacity to manage a diverse range of input variations.

\begin{table*}[t!]
    \centering
    \caption{Comparison of mean WS metric across 4 quartiles of different intensity ranges. 
    ExpertSim offers better WS across all quartiles when compared with standard GAN and SDI-GAN with intensity regularization~(IR) and auxiliary regressor~(AR). Boldface indicates the best-performing result, while underlining denotes the second-best.}
    \vspace{0.3cm}
    \setlength{\tabcolsep}{5pt} %
    \small
    \begin{tabular}{l@{\hskip 1cm}cccccccc}
        \toprule
        \textbf{Model} & \multicolumn{4}{c}{\textbf{Proton}} & \multicolumn{4}{c}{\textbf{Neutron}} \\
        \cmidrule(lr){2-5} \cmidrule(lr){6-9}
        & \textbf{1st Qrt} & \textbf{2nd Qrt} & \textbf{3rd Qrt} & \textbf{4th Qrt} 
        & \textbf{1st Qrt} & \textbf{2nd Qrt} & \textbf{3rd Qrt} & \textbf{4th Qrt} \\
        \midrule
        GAN & 4.07 & 4.20 & 3.76 & 18.13  & 4.82 & 5.37 & 7.75 & 10.39 \\
        SDI-GAN & 4.37 & 5.07 & 4.78 & 15.08 & 4.67 & 5.18 & 7.52 & 9.67  \\
        SDI-GAN + IR & \underline{3.33} & \underline{3.42} & \underline{3.47} & \underline{9.92} & \underline{4.48} & \underline{4.91} & 6.69 & 8.29 \\
        SDI-GAN + IR + AR & 3.52 & 4.20 & 4.48 & 12.72 & 4.49 & 5.02 & \underline{6.34} & \underline{7.05} \\
        \textbf{\ours{}} & \textbf{1.06} & \textbf{1.16} & \textbf{1.27} & \textbf{7.06} & \textbf{2.20} & \textbf{4.02} & \textbf{5.62} & \textbf{5.69} \\
        \bottomrule
    \end{tabular}
    \vspace{0.2cm}
    \label{tab:quart}
\end{table*}

\begin{table}[t!]
    \centering
    \vspace{0.2cm}
    \caption{Comparison of mean WS metric across five runs. 
    Our method outperforms standard GAN and SDI-GAN with intensity regularization~(IR) and auxiliary regressor~(AR). 
    We follow ~\citet{Dubinski2023MachineLM} and train our method on data generated for the ZDC using a Monte Carlo-based simulation, which can be treated as the upper bound for our solution. For the Monte Carlo simulation, we calculate the Wasserstein metrics between two runs.
    }
    \vspace{0.3cm}
    \small
    \begin{tabular}{lcccc}
    \toprule
    \textbf{Model} & \multicolumn{2}{c}{\textbf{Proton}} & \multicolumn{2}{c}{\textbf{Neutron}} \\
    \cmidrule(lr){2-3} \cmidrule(lr){4-5}
    & \textbf{WS↓} & \textbf{Std. Dev.} & \textbf{WS↓} & \textbf{Std. Dev.} \\
    \midrule
    GAN & 2.47 & 0.010 & 2.42 & 0.041 \\
    SDI-GAN & 2.35 & 0.027 & 2.31 & 0.063 \\
    SDI-GAN + IR & 2.29 & \underline{0.009} & 2.04 & \underline{0.018} \\
    SDI-GAN + IR + AR & \underline{2.07} & 0.025 & \underline{1.89} & 0.029 \\
    \textbf{\ours{}} & \textbf{1.59} & \textbf{0.008} & \textbf{1.34} & \textbf{0.010} \\
    \midrule
    Monte-Carlo & 0.47 & 0.033 & 0.34 & 0.014 \\
    \bottomrule
    \label{tab:ws_comparison}
    \end{tabular}
    \vspace{0.2cm}
\end{table}

\subsubsection{Final Router Training Objective}

The final router training objective \( L_{\text{router}} \) combines the sum of the generator-discriminator losses across all experts with regularization terms for expert utilization and differentiation:
\begin{equation}
    L_{\text{router}} = \sum_{i=1}^{N} L(G_i, D_i) + \lambda_{\text{util}} L_{\text{util}} + \lambda_{\text{diff}} L_{\text{diff}},
\end{equation}
where \( L(G_i, D_i) \) represents the generator-discriminator loss for each expert \( i \), \( \lambda_{\text{util}} \) is a regularization weight for the expert utilization entropy loss \( L_{\text{util}} \) which promotes balanced routing, and \( \lambda_{\text{diff}} \) is a regularization weight for the expert differentiation loss \( L_{\text{diff}} \) which encourages diversity among expert outputs.

\section{Experiments}

The proton and neutron datasets employed in this work consist of 344 thousand and 400 thousand samples, respectively, with the validation method incorporating an 80:20 train-test split ratio. Our evaluation methodology is founded on the analysis of five distinct channels outlined in the calorimeter's specifications \citep{alice-tech-report}. Those five channel values correspond to the number of photons collected by each of the five distinct photomultipliers (PMTc1, PMT1-4) in ZP and ZN. We employ the standard 1st Wasserstein distance metric \citep{ws-dist} to assess the fidelity of the simulations across all channels by comparing the generations to the ground truth simulations from the test set. The code to reproduce our experiments is available in the repository\footnote{\url{https://github.com/patrick-bedkowski/expertsim-mix-of-generative-experts}}. %

\subsection{Results}

\begin{figure*}[t!]
    \centering
    \includegraphics[width=0.33\textwidth, trim={1.5cm 0.5cm 1.5cm 0.25cm}, clip]{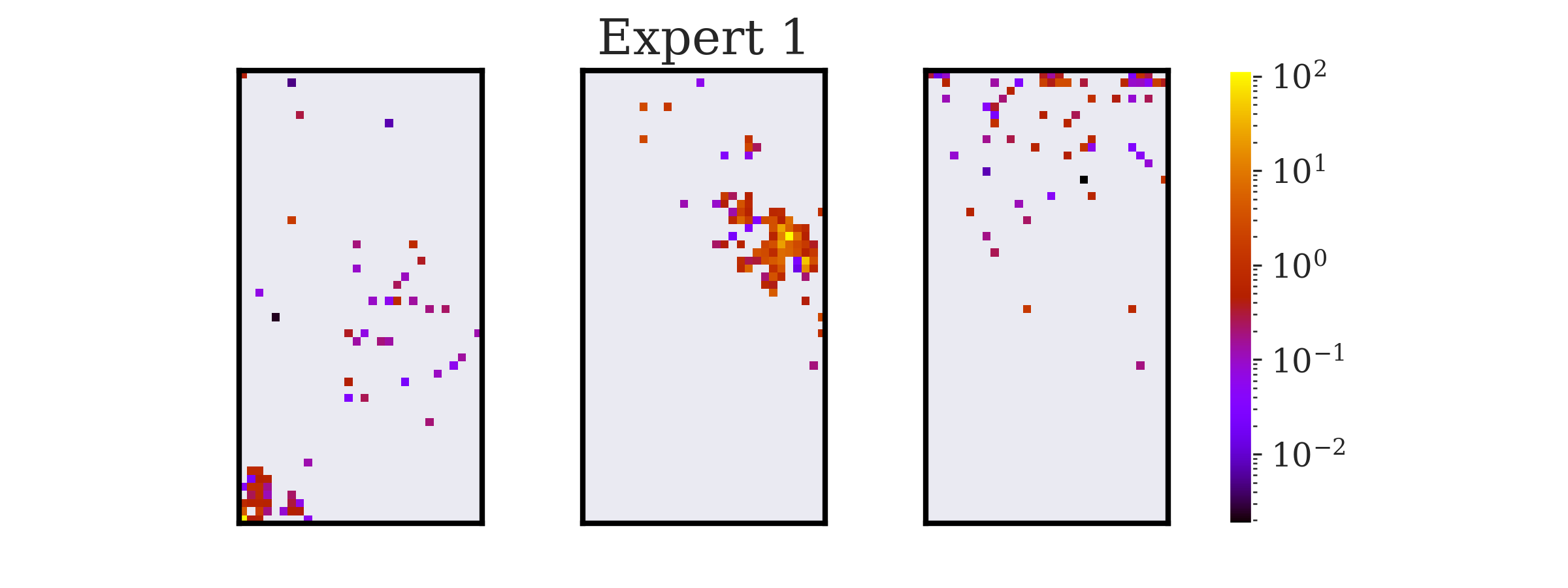} 
    \includegraphics[width=0.33\textwidth, trim={1.5cm 0.5cm 1.5cm 0.25cm}, clip]{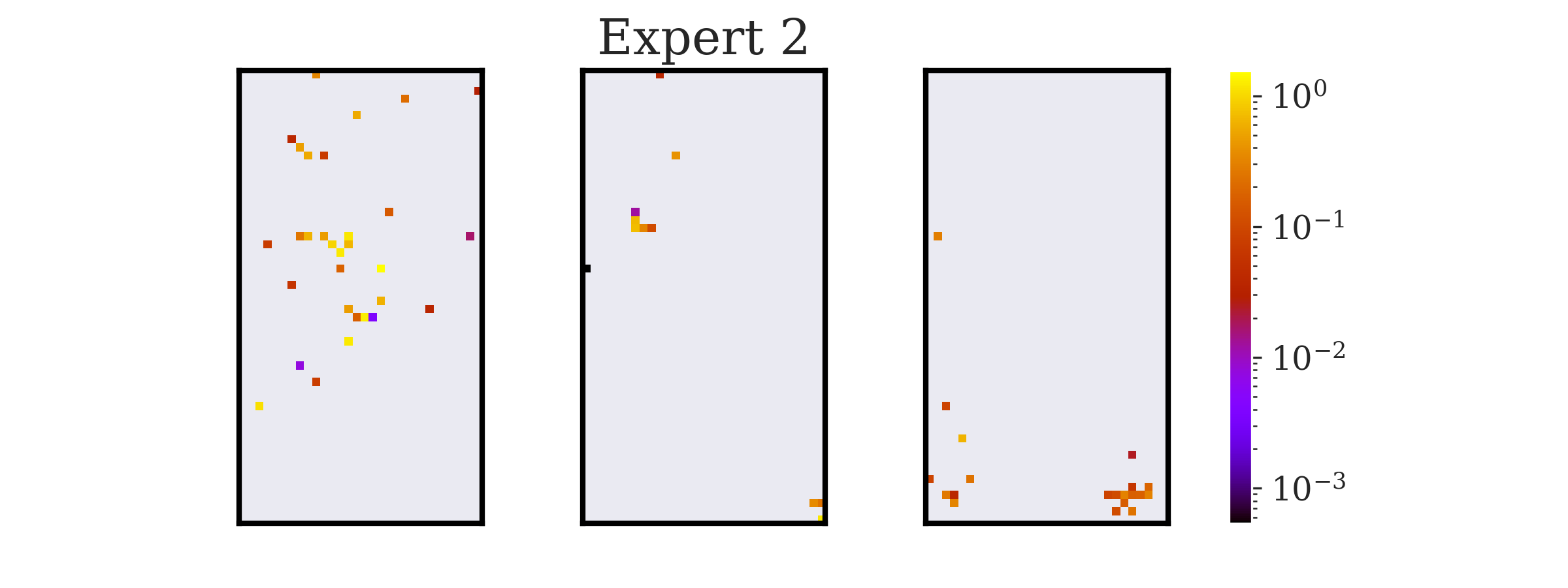} 
    \includegraphics[width=0.33\textwidth, trim={1.5cm 0.5cm 1.5cm 0.25cm}, clip]{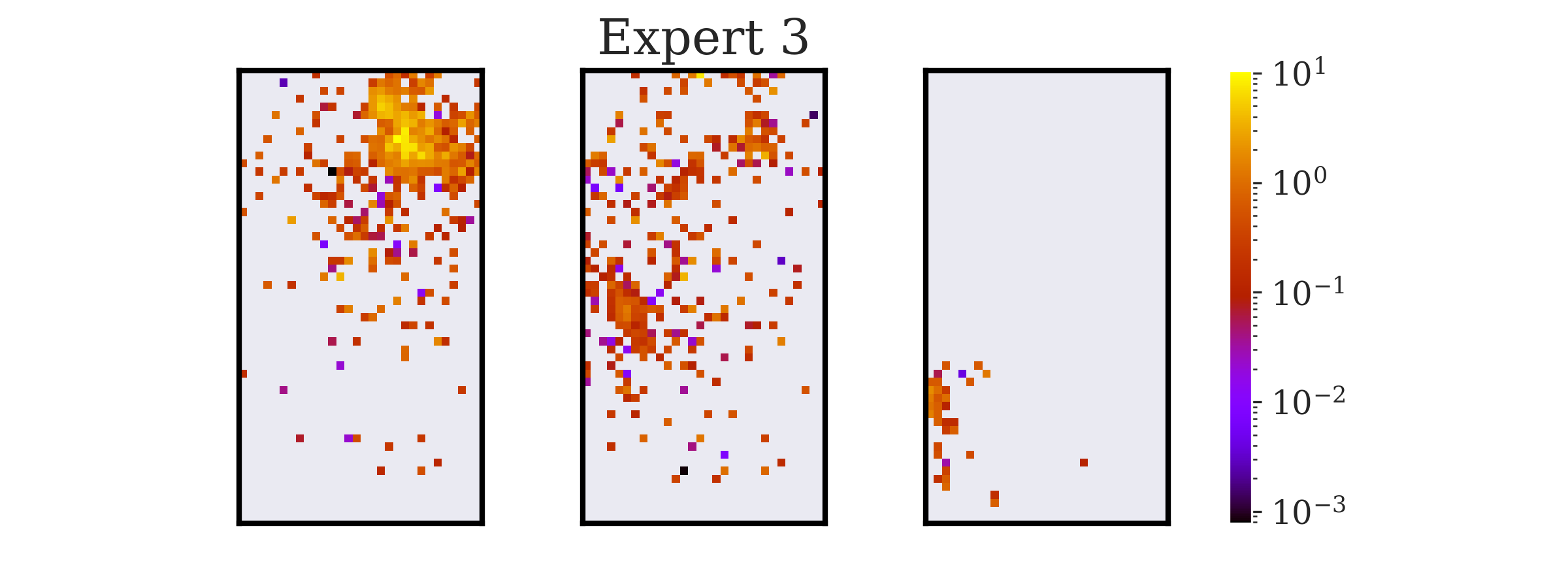}

    \label{fig:expert_examp}
\end{figure*}

\begin{figure*}[t!]
    \centering
    \includegraphics[width=0.33\textwidth, trim={1cm 0.5cm 1cm 0.5cm}, clip]{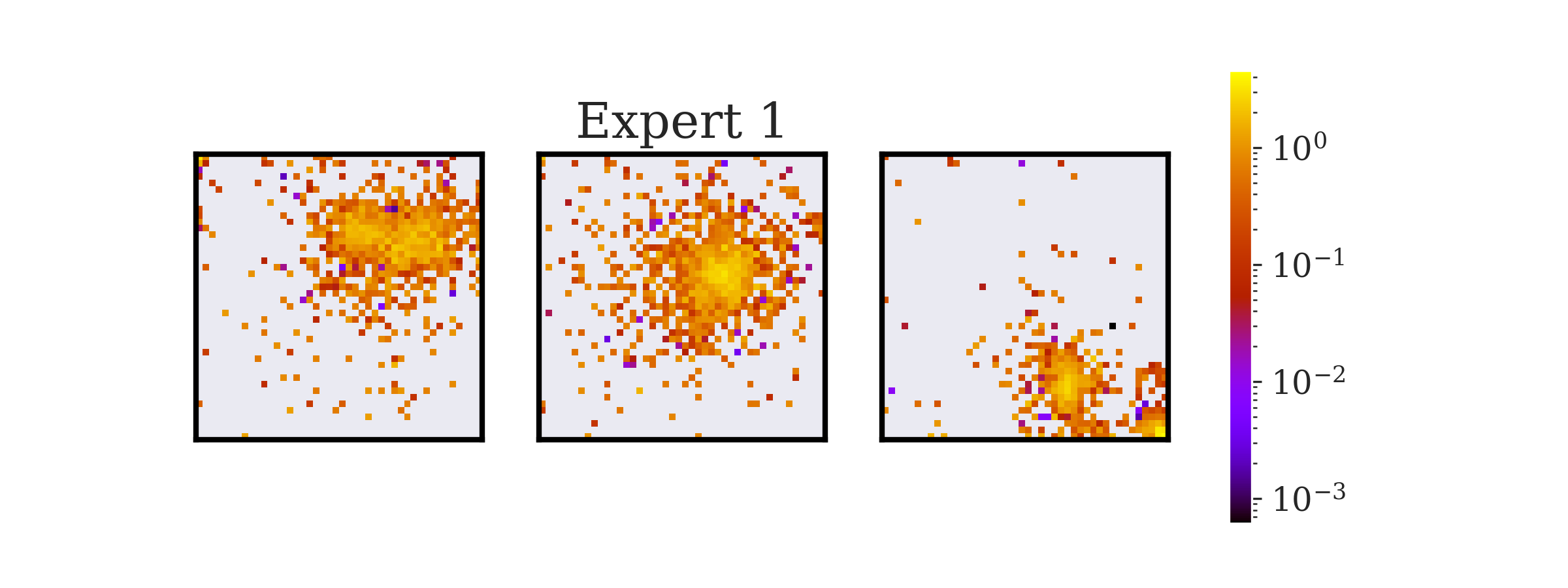} 
    \includegraphics[width=0.33\textwidth, trim={1cm 0.5cm 1cm 0.5cm}, clip]{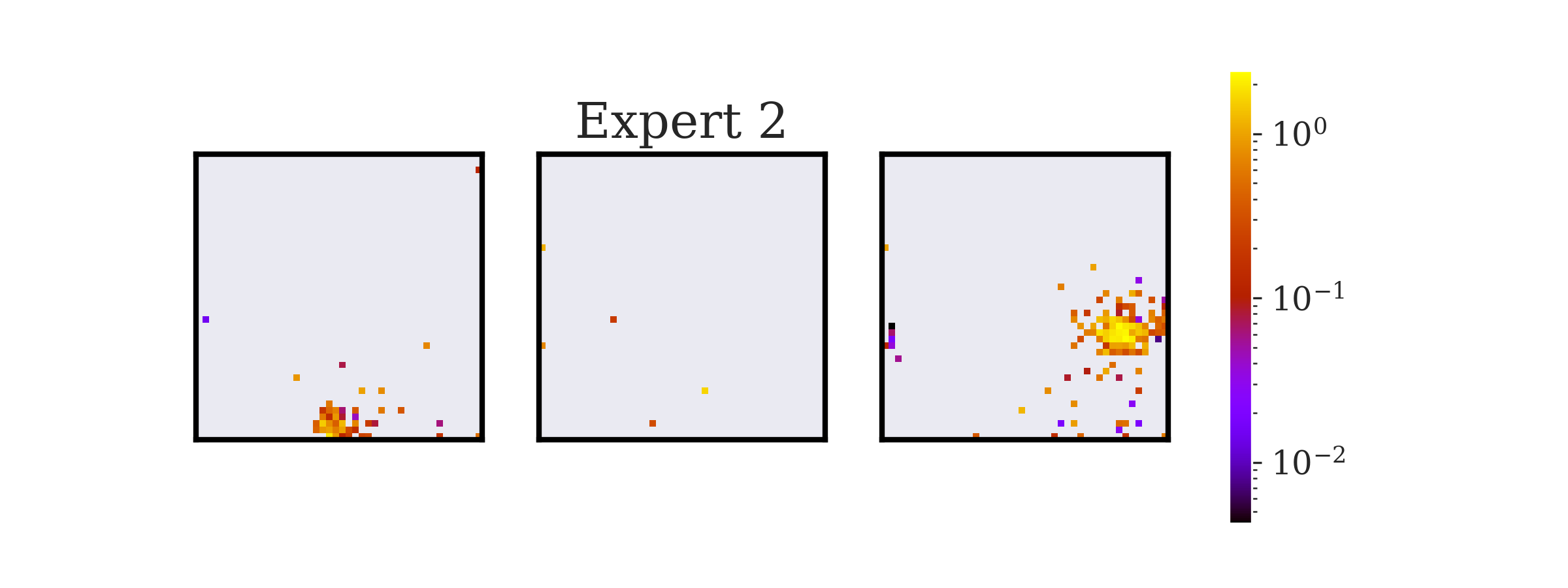} 
    \includegraphics[width=0.33\textwidth, trim={1cm 0.5cm 1cm 0.5cm}, clip]{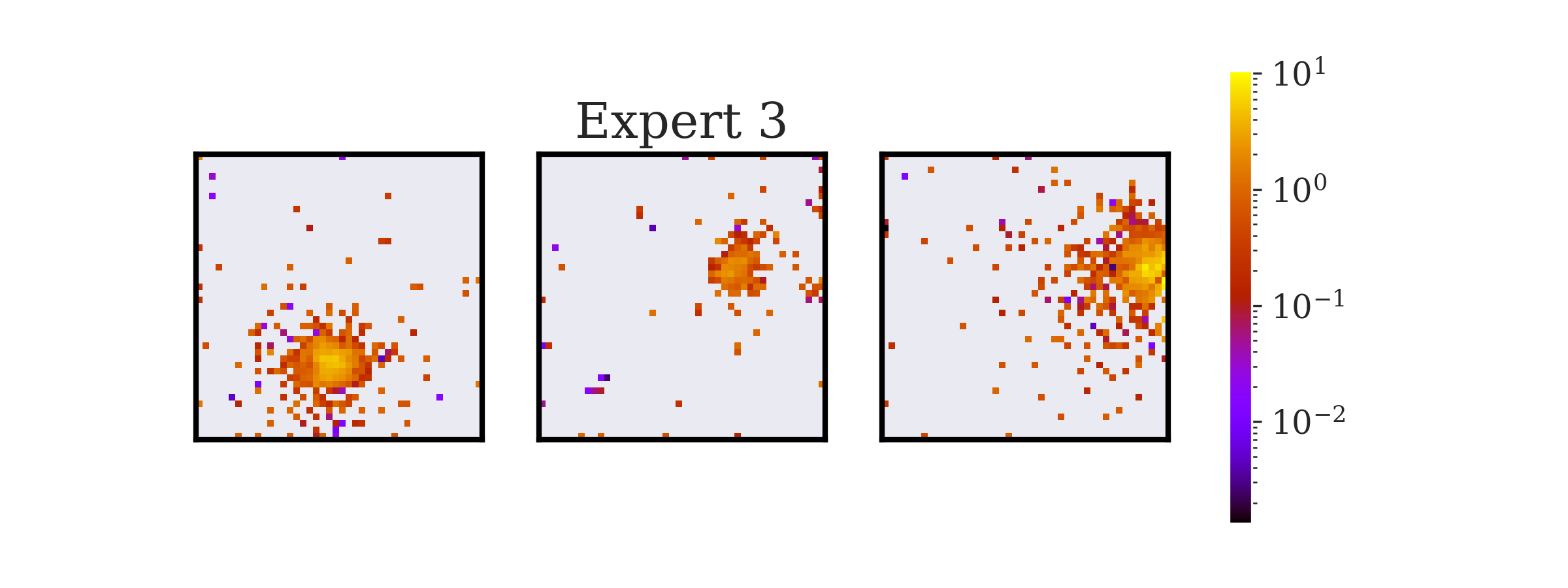} 
\vspace{0.2cm}
  \caption{Examples of images generated by each expert. Top: ZP, Bottom: ZN. Please note the differing color bar scale for better visibility.}
    \label{fig:expert_examp}
    \vspace{0.2cm}
\end{figure*}

We further compare the ExpertSim model with previous single-generator methods, GAN and SDI-GAN variants, using the Wasserstein distance metric (WS). As shown in Tab.~\ref{tab:ws_comparison}, incorporating diversity regularization in SDI-GAN improves the metric compared to the standard GAN. Further regularization on intensity and the use of an auxiliary regressor provide the best performance among the single-generator models. However, ExpertSim provides a significant reduction in WS distance, achieving the best generation quality across all approaches. The inference time on GPU hardware is nearly identical to that of single-generator models, with the overhead introduced by the router network equal to 2\%. Moreover, we achieve a speedup of more than an order of magnitude compared to the Monte Carlo simulation.
We include more details on this comparison in Tab.~\ref{tab:inference_time}.

\begin{figure*}[h!]
    \vspace{0.6cm}
    \includegraphics[width=1\textwidth]{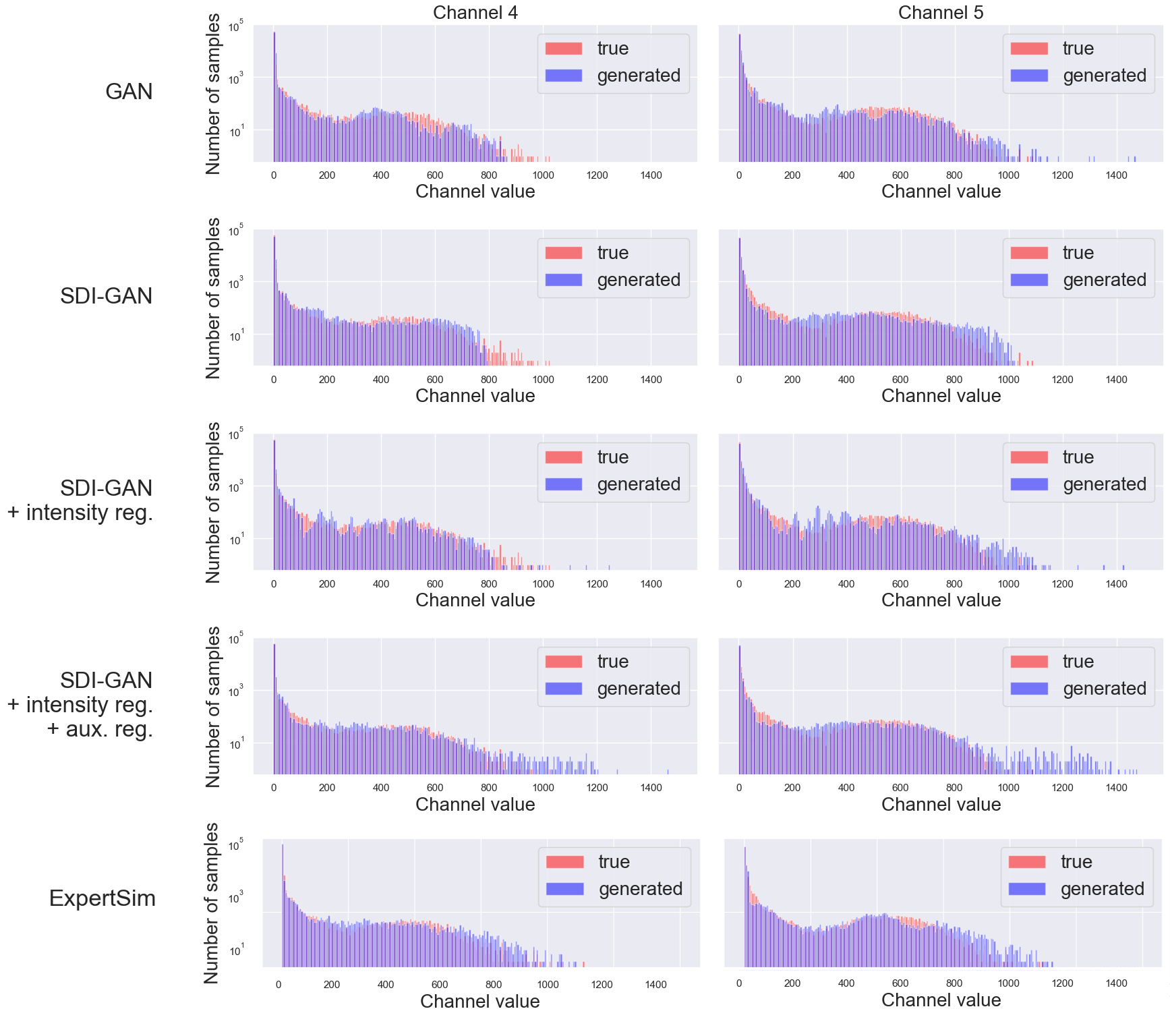}
    \caption{Histograms of true and generated distributions of ZP channel values.}
    \label{fig:histograms_of_channels}
    \vspace{0.2cm}
\end{figure*}

In Fig.~\ref{fig:histograms_of_channels}, we present the visual results for channels 4 and 5, as they contain the majority of the relevant information. The GAN and SDI-GAN models have visible problems with underproducing high-energy responses. The implementation of additional regularization and auxiliary regressor positively influences better alignment to true distribution but tends to oversample the high-energy responses. ExpertSim model better aligns with the true distribution for both low and high-value channels. 

We further evaluate the models' performance across different intensity ranges by dividing the entire range of sorted ascending intensities from the original dataset into four quartiles. For each quartile, we calculate the WS metric to assess how well the models perform within these specific intensity ranges and report the results in Tab.~\ref{tab:quart}. Our ExpertSim method outperforms all other models by achieving the lowest WS distance across all intensity ranges.

\subsection{Experts specialization}

To highlight the main benefit of our Mixture-of-Experts architecture, we propose to study the specialization of each individual expert. To this end, we calculate the mean intensity of responses generated by each expert and report their results in Tab.~\ref{tab:expers_spec}.
Moreover, in Fig.~\ref{fig:expert_examp}, we present exemplar outputs generated by each of the three experts. %

As evidenced by the higher WS distance in Tab.~\ref{tab:ws_comparison}, single generator methods are unable to capture the whole distribution of the training data. In contrast, in ExpertSim, specific GAN Experts learn different data subsets as shown in Tab.~\ref{tab:expers_spec}. Each expert specializes in different types of detector responses, which allows us to improve the final performance of the model. This leads to improved performance in our domain, where the ZDC produces responses of distinct types.

\begin{table}[t]
    \centering
    \small
    \vspace{0.2cm}    
    \caption{Mean intensities of responses generated by each expert. Our experts specialize in generating samples with different energies.}
    \vspace{0.3cm}
    \begin{tabular}{lcccc}
    \toprule
    \textbf{Model} & \multicolumn{2}{c}{\textbf{Proton}} & \multicolumn{2}{c}{\textbf{Neutron}} \\
    \cmidrule(lr){2-3} \cmidrule(lr){4-5}
    & \textbf{Mean} & \textbf{Std. Dev.} & \textbf{Mean} & \textbf{Std. Dev.} \\
    \midrule
    Expert 1 & 119 & 340 & 108 & 100 \\
    Expert 2 & 53 & 204 & 82 & 85 \\
    Expert 3 & 70 & 230 & 125 & 113 \\
    \bottomrule
    \label{tab:expers_spec}
    \end{tabular}
\end{table}

\subsection{Ablation study}
\label{sec:ablation}
\reb{In our design, we maintain a balance between methodological complexity and simulation fidelity. While our framework introduces additional hyperparameters compared to single-model approaches, each component contributes meaningfully to the overall performance. 
The hyperparameters related to individual experts follow values established in prior work \citep{bedkowski2024deep}, with our tuning efforts focused specifically on the router network parameters ($\lambda_{\text{util}}$, $\lambda_{\text{diff}}$) that govern expert specialization and utilization.
}

\begin{table}[t]
\centering
\small
\caption{Effect of the number of experts on the final performance of the generative model. Using 3 experts achieves the best performance for both proton and neutron responses.}
\vspace{0.2cm}
\scalebox{0.9}{%
\begin{tabular}{ccc}
\toprule
\textbf{Number of Experts} & \textbf{Proton WS$\downarrow$} & \textbf{Neutron WS$\downarrow$} \\
\midrule
2 & 2.71 & 1.76 \\
\textbf{3} & \textbf{1.59} & \textbf{1.34} \\
4 & 3.26 & 1.63 \\
5 & 3.07 & 3.12 \\
\bottomrule
\end{tabular}%
}
\vspace{0.2cm}
\label{tab:experts}
\end{table}

\begin{table}[t]
\centering
\small
\caption{Performance (WS) with varying router hyperparameter values. 
Our method remains robust across a wide range of $\lambda_{\text{util}}$ and $\lambda_{\text{diff}}$ values for both proton and neutron simulations.}
\vspace{0.2cm}
\scalebox{0.9}{%
\begin{tabular}{ccccccl}
\toprule
\multicolumn{2}{c}{\textbf{Hyperparameters}} & \multicolumn{2}{c}{\textbf{WS$\downarrow$}} & \multicolumn{2}{c}{\textbf{Comment}} \\
\cmidrule(r){1-2} \cmidrule(lr){3-4}
$\lambda_{\text{util}}$ & $\lambda_{\text{diff}}$ & \textbf{Proton} & \textbf{Neutron} & \multicolumn{2}{l}{} \\
\midrule
0.01  & 0.0001   & \textbf{1.59} & \textbf{1.34} & \multicolumn{2}{l}{Default} \\
0.1   & 0.0001   & 1.86          & 4.14          & \multicolumn{2}{l}{Higher $\lambda_{\text{util}}$} \\
0.001 & 0.00001  & 2.34          & 2.52          & \multicolumn{2}{l}{Lower $\lambda_{\text{util}}$} \\
0.01  & 0.001    & 1.92          & 1.91          & \multicolumn{2}{l}{Higher $\lambda_{\text{diff}}$} \\
0.01  & 0.000001 & 1.78          & 1.74          & \multicolumn{2}{l}{Lower $\lambda_{\text{diff}}$} \\
0.01  & 0        & 6.80          & 8.01          & \multicolumn{2}{l}{$L_{\text{diff}}$ disabled} \\
0     & 0.0001   & 2.10          & 1.89          & \multicolumn{2}{l}{$L_{\text{util}}$ disabled} \\
\bottomrule
\end{tabular}%
}
\vspace{0.2cm}
\label{tab:hyperparams}
\end{table}

The number of experts we use (3) is motivated by the characteristics of our problem, as the calorimeters (both ZP and ZN) generally produce: 1) low-intensity, dispersed responses, 2) medium-intensity responses, and 3) high-intensity, focused responses.
To confirm our insight experimentally, we run an additional ablation study with a~different number of experts. As presented in Tab.~\ref{tab:experts}, using 3 experts outperforms other approaches, which confirms our assumptions. %

$\lambda_{\text{util}}$ and $\lambda_{\text{diff}}$ are key components for our routing, where $L_{\text{diff}}$ enforces expert specialization, and $L_{\text{util}}$ promotes balanced distribution of tasks. We study the effect of their values in Tab.~\ref{tab:hyperparams}. As visible, disabling $L_{\text{diff}}$ leads to poor specialization among agents, while without $L_{\text{util}}$, our routing collapses. However, apart from those extreme cases, our method is robust to the exact values of $\lambda_{\text{util}}$ and $\lambda_{\text{diff}}$. For the remaining hyperparameters specific to individual generative agents, we select values provided by \citet{sdigan2023}.

\reb{While we strive for simplicity in our approach, the number of hyperparameters in \ours{} is driven by our goal of achieving optimal simulation fidelity. As demonstrated in Tab.~\ref{tab:ws_comparison}, each additional component (diversity regularization, intensity regularization, auxiliary regression) progressively improves SDI-GAN's performance, with our ExpertSim ultimately surpassing all variants. These performance improvements justify the additional hyperparameters. As we build on top of \citet{bedkowski2024deep}, we maintain consistency by reusing their hyperparameter values for individual experts, focusing our tuning efforts only on router-specific parameters ($\lambda_{\text{util}}$, $\lambda_{\text{diff}}$) through a compact logarithmic grid search. To minimize tuning complexity, we share all hyperparameter values across experts. As Tab.~\ref{tab:hyperparams} demonstrates, our framework exhibits considerable robustness to parameter selection within reasonable ranges.}

\begin{table}[t!]
\centering
\small
\caption{Inference time across different numbers of experts. MC refers to a Monte Carlo ensemble.}
\vspace{0.2cm}
\scalebox{0.9}{%
\begin{tabular}{@{}ccc@{}}
\toprule
\textbf{Number of Experts} & \textbf{CPU Time [s]} & \textbf{GPU Time [s]} \\ 
\midrule
1 & 273 & 3.20 \\
2 & 315 & 3.21 \\
3 & 314 & 3.27 \\
4 & 324 & 3.44 \\
5 & 333 & 3.55 \\
\midrule
Monte-Carlo & 11172 & -- \\
\bottomrule
\end{tabular}%
}
\vspace{0.2cm}
\label{tab:inference_time}
\end{table}

\subsection{Inference time and scalability.}
The ALICE experiment at CERN relies on CPU-based infrastructure, including Monte Carlo simulations that run on the CERN computational grid. To ensure a fair comparison of simulation speed across methods, we benchmark all approaches on a CPU to reflect this target deployment environment. Tab.~\ref{tab:inference_time} presents the inference time required to simulate 10,000 responses using our method with varying numbers of experts, as well as a baseline Monte Carlo (MC) ensemble. All CPU benchmarks are conducted on an Intel Core i7-9800X. Given that our method can also be efficiently executed on GPUs, we additionally report inference times with a single Nvidia A5000 GPU.

We can observe that there is a small overhead in the inference time as we increase the number of experts. This can be attributed to the fact that with each added expert, we additionally adapt the shared part of the network to accommodate the extended overall architecture.  %
Nevertheless, the resulting performance is an order of magnitude faster than the standard Monte-Carlo approach.

\section{Conclusions}

In this work, we applied and expanded generative modeling techniques to simulate the responses of the Zero Degree Calorimeter within the ALICE experiment at CERN. Using Mixture-of-Generative-Experts approach, we leveraged multiple GAN-based experts to improve the fidelity of our simulation framework. 

We incorporated SDI-GAN to boost the diversity of generated samples, and we implemented a tailored intensity regularization method to minimize differences between real and generated calorimeter responses. Additionally, the use of an auxiliary regressor enabled more accurate spatial feature learning, further enhancing the simulation's fidelity.

By training a router network that assigns samples based on physical characteristics, our approach not only improved the quality of generated data but also demonstrated the potential for efficient simulation of high-energy physics experiments. The results indicate that our MoE-based simulation method offers an alternative to traditional Monte-Carlo methods, outperforming approaches based on a single generative model.

\section*{Acknowledgements}
This work is supported by National Centre of Science (NCP, Poland) Grants No. 2022/45/B/ST6/02817, 2024/53/N/ST6/03078, 2023/51/B/ST6/03004 and 2023/51/D/ST6/02846. 
This work was supported by Horizon Europe Programme under GA no. 101120237, project "ELIAS: European Lighthouse of AI for Sustainability".
We gratefully acknowledge Polish high-performance computing infrastructure PLGrid (HPC Center: ACK Cyfronet AGH) for providing computer facilities and support within computational grant no. PLG/2025/018551.

\clearpage
\printbibliography

\end{document}